\documentclass[letterpaper, 10 pt, conference]{ieeeconf}  % Comment this line out if you need a4paper
\let\labelindent\relax

\IEEEoverridecommandlockouts                              % This command is only needed if 
\usepackage{amssymb}  % assumes amsmath package installed
\usepackage{multirow}

\usepackage{marvosym}
\usepackage{makecell}
\usepackage{url}
\usepackage{enumitem}
\usepackage{tabularx} % Import tabularx for flexible-width tables
\usepackage{array}    % For better table control

\usepackage{graphics} % for pdf, bitmapped graphics files
\usepackage{epsfig} % for postscript graphics files
\usepackage{times} % assumes new font selection scheme installed
\usepackage{stfloats}
\usepackage{cite}

\usepackage{tcolorbox}

\title{\LARGE \bf
A Comprehensive LLM-powered Framework for\\ Driving Intelligence Evaluation
}

\author{Shanhe You, Xuewen Luo, Xinhe Liang, Jiashu Yu, Chen Zheng and Jiangtao Gong\textsuperscript{\Letter}% <-this % stops a space
\thanks{The authors are with the Institute for AI Industry Research, Tsinghua University, Beijing, China. Corresponding Email:
        {\tt\small gongjiangtao@air.tsinghua.edu.cn}}
}

\begin{document}

\maketitle
\thispagestyle{empty}
\pagestyle{empty}

%%%%%%%%%%%%%%%%%%%%%%%%%%%%%%%%%%%%%%%%%%%%%%%%%%%%%%%%%%%%%%%%%%%%%%%%%%%%%%%%
\begin{abstract}

Evaluation methods for autonomous driving are crucial for algorithm optimization. However, due to the complexity of driving intelligence, there is currently no comprehensive evaluation method for the level of autonomous driving intelligence. In this paper, we propose an evaluation framework for driving behavior intelligence in complex traffic environments, aiming to fill this gap. We constructed a natural language evaluation dataset of human professional drivers and passengers through naturalistic driving experiments and post-driving behavior evaluation interviews. Based on this dataset, we developed an LLM-powered driving evaluation framework. The effectiveness of this framework was validated through simulated experiments in the CARLA urban traffic simulator and further corroborated by human assessment. Our research provides valuable insights for evaluating and designing more intelligent, human-like autonomous driving agents. The implementation details of the framework\footnote{\url{https://github.com/AIR-DISCOVER/Driving-Intellenge-Evaluation-Framework}} and detailed information about the dataset\footnote{\url{https://github.com/AIR-DISCOVER/Driving-Evaluation-Dataset}} can be found at the provided links.
\end{abstract}

%%%%%%%%%%%%%%%%%%%%%%%%%%%%%%%%%%%%%%%%%%%%%%%%%%%%%%%%%%%%%%%%%%%%%%%%%%%%%%%%
\section{INTRODUCTION}

The field of autonomous driving evaluation has evolved significantly, transitioning from traditional, single-layer mechanical metrics to more comprehensive, multi-dimensional evaluation frameworks\cite{hou2021visual,guo2018safe,xu2022safebench}. Early approaches relied heavily on mathematical models to simulate driving scenarios, but these have now expanded to incorporate AI-annotated machine learning techniques\cite{10.1145/3540250.3549111,erdougan2021evaluation,wang2019method}. Despite these advances, current evaluation metrics remain rigid, lacking the flexibility necessary to adapt to the complexities of real-world driving\cite{gao2021quantitative}. A critical shortcoming of these traditional methods is their inability to effectively assess the intelligence of autonomous systems, as they fail to align with the cognitive processes humans use when evaluating driving performance.

Large language models (LLMs) have demonstrated superior capabilities in complex reasoning and knowledge application \cite{NEURIPS2022_9d560961}, opening new possibilities in autonomous driving for path planning \cite{seff_motionlm_2023,mao_gpt-driver_2023,sharan_llm-assist_2023,sima_drivelm_2024} and decision interpretability \cite{zablocki_explainability_2022,yuan_rag-driver_2024,chen_driving_2023}.
LLMs offer potential solutions to current evaluation limitations by mirroring human thought processes \cite{chang2024survey}. Effective evaluation requires careful selection of key factors as prompts \cite{li2024automated}, incorporating both fundamental driving elements and higher-level reasoning.
Besides, LLM-as-a-judge approaches \cite{zheng2024judging} have shown promise in generating neutral, interpretable evaluations across various domains \cite{chang2024survey,gao2024llm}, with recent advancements in unsupervised evaluation \cite{yuan2024self, wu2024meta} and domain knowledge integration \cite{pan2024human,li2024automated,tang_testing_2024}.
However, LLMs still lack sufficient understanding of complex, domain-specific knowledge, limiting their ability to fully evaluate critical driving information.

\begin{figure}
    \centering
    \includegraphics[width=0.93\linewidth]{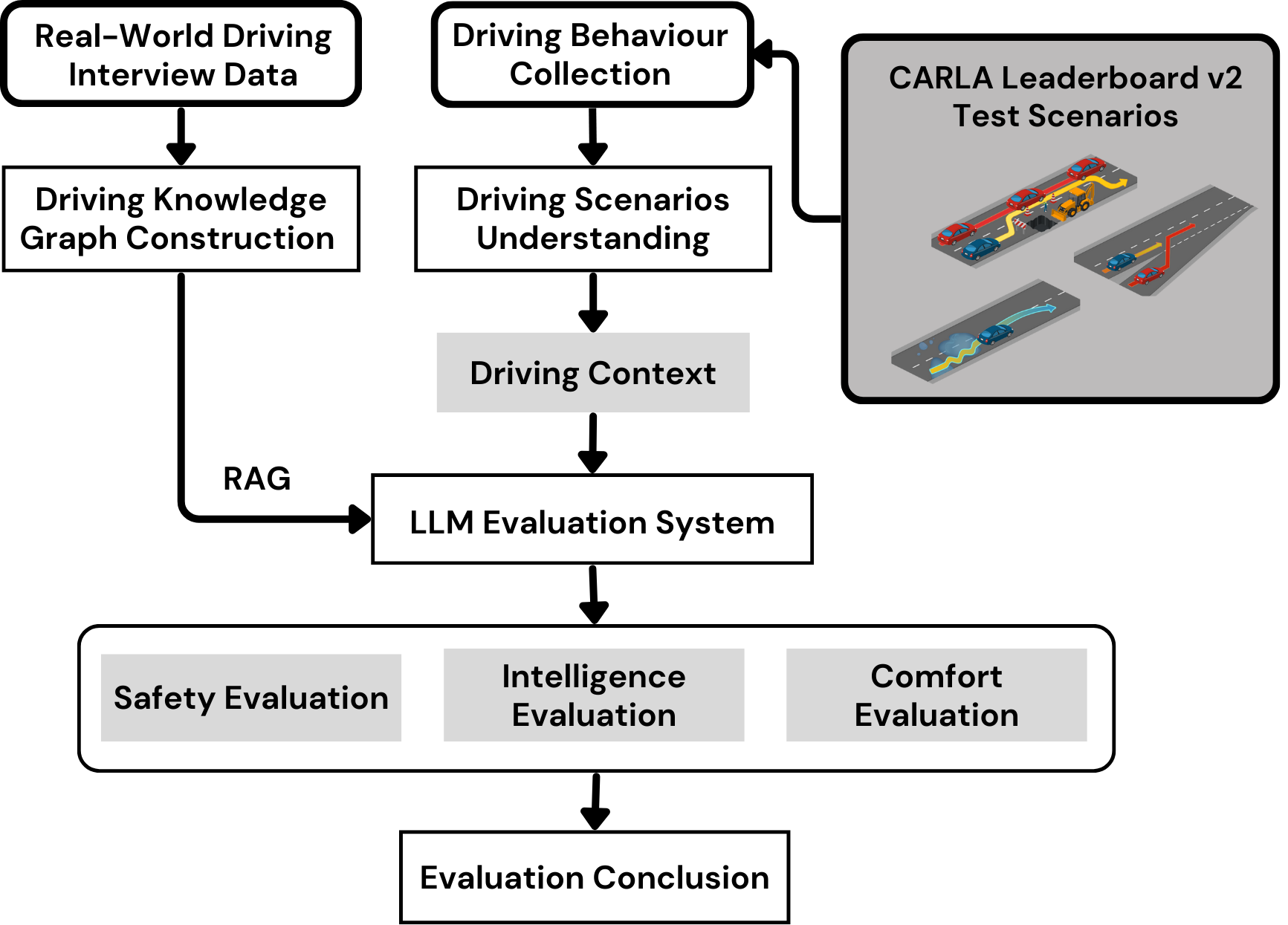}
    \caption{A comprehensive framework for evaluating driving intelligence using LLMs. We use real-world driving interview data to construct a driving evaluation knowledge graph, which, together with driving behaviors collected from the simulator, generates driving context. Then, it generates assessments on safety, intelligence, and comfort, leading to a comprehensive evaluation of the driving performance.}
    \label{fig:main_figure}
\end{figure}

To this end, we propose a novel LLM-based evaluation framework to assess driving behavior intelligence in complex traffic environments, addressing the gap in comprehensive evaluation methods. As shown in Fig.~\ref{fig:main_figure}, we developed this framework by first constructing a natural language evaluation dataset from human professional drivers and passengers through naturalistic driving experiments and post-driving behavior evaluation interviews. We then developed an LLM-powered driving evaluation framework. To ensure its effectiveness, we validated the framework through simulated experiments in the CARLA urban traffic simulator and further corroborated the results with human assessment. 
The contributions of this paper can be summarized as follows:
\begin{enumerate}
    \item We collected and constructed a driving evaluation dataset through naturalistic driving experiments and post-experiment interviews with drivers and passengers.
    \item Based on in-depth analysis of the driving evaluation data, we developed a comprehensive evaluation framework for assessing autonomous driving behaviors.
    \item We implemented this framework using LLM and validated it through human assessment with data collected in the driving simulator.
\end{enumerate}
\section{Related Work}
\textbf{Autonomous Driving Evaluation:} Evaluation metrics in autonomous driving studies encompass basic\cite{gao2021quantitative, dosovitskiy2017carla}, etiquette\cite{xu2022safebench}, comfortableness\cite{hou2021visual,wang2019method,fu2024limsim++,roshdi2020unified,xu2022safebench}, risk\cite{guo2018safe,erdougan2021evaluation,chia2022risk}, and trust\cite{azevedo-sa_real-time_2021,takeover,9802974,kuhnlenz_study_2023} categories. Kim et al.\cite{takeover} investigated the relationship between driving scenarios and take-over events, defining DRI for safety evaluation. Kühnlenz et al.\cite{kuhnlenz_study_2023} proposed that users' perceptions of system intelligence are influenced by prior information and continuous self-explanations, enhancing transparency and trust. However, intelligent assessment based on specific driving scenarios remains at the analytical level, and human-like evaluation methods reflecting driving philosophy are needed. The field lacks comprehensive methods that integrate scenario-specific assessments with human-like decision-making processes.

\textbf{LLM in Driving:} LLMs have shown promise in driving by integrating common-sense knowledge\cite{sharan_llm-assist_2023} with machine behavior\cite{seff_motionlm_2023}. They combine objective domain knowledge (e.g., traffic rules\cite{li_large_2024}) with subjective human-generated data\cite{han_dme-driver_2024,yuan_rag-driver_2024,chen_driving_2023,xu_drivegpt4_2024}, enabling nuanced understanding of human-like driving behavior. LLMs effectively manage motion planning tasks\cite{tian_drivevlm_2024,huang_drivlme_2024,seff_motionlm_2023,sharan_llm-assist_2023,wang_drivemlm_2023} through techniques like temporally causal decoders\cite{seff_motionlm_2023}. The prompting-reasoning-finetuning strategy\cite{mao_gpt-driver_2023,xu_drivegpt4_2024} enhances decision-making capabilities. LLMs offer explainability\cite{zablocki_explainability_2022} by providing transparent reasoning for driving decisions, promoting human-aligned\cite{wen_dilu_2024,zhang_feedback-guided_nodate,sun_optimizing_2024,yuan_rag-driver_2024,pan_vlp_nodate}, knowledge-driven approaches\cite{wen_dilu_2024,ananthajothi_advancing_2024}. However, they lack embodied intelligence, focusing on general knowledge rather than specific driving expertise, which limits their understanding of complex driving scenarios.

\textbf{LLM-as-a-judge:} LLMs generate neutral, interpretable, and scalable evaluations across diverse domains\cite{zheng2024judging,chang2024survey,gao2024llm}. Recent research has employed advanced methodologies like META-REWARDING\cite{yuan2024self, wu2024meta} for unsupervised evaluation, enabling LLMs to autonomously refine their judgment capabilities. LLMs can incorporate domain knowledge\cite{pan2024human} and assess corner cases\cite{li2024automated} or driving skills\cite{tang_testing_2024}. They provide QA-based driving behavior explanations\cite{sima_drivelm_2024,shao_lmdrive_2023,han_dme-driver_2024,chen_driving_2023,cui_personalized_2024,wang_omnidrive_2024,2023LingoQA}, enhancing interpretability. However, LLMs often lack sufficient understanding of complex, domain-specific knowledge, especially in automated driving. This deficit hampers their ability to fully and accurately evaluate critical driving information, particularly in nuanced or rapidly changing scenarios.

\begin{figure*}
    \centering
    \includegraphics[width=0.9\textwidth]{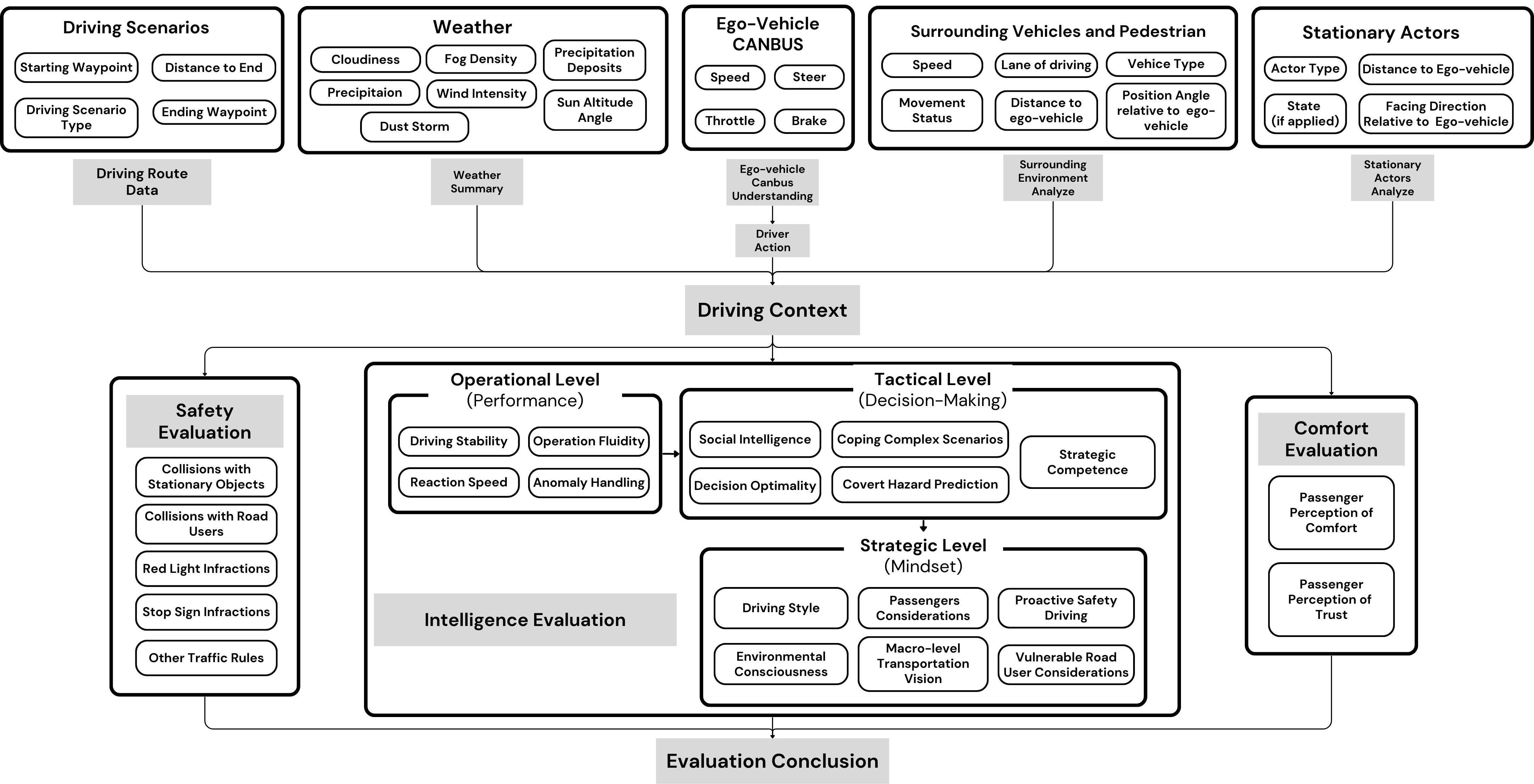}
    \caption{Framework Structure for Driving Intelligence Evaluation}
    \label{fig:framework}
\end{figure*}

\section{DRIVING-EVALUATION DATASET}

\subsection{Driving Experiment and Interview Data Acquisition}

To collect real-world driving evaluation data from humans, we designed and implemented real-world driving experiments. In the experiment, we conducted in-depth interviews with every driver and passenger to understand their evaluations of this driving performance comprehensively. A total of 24 drivers and 48 passengers were invited to participate in this experiment. These participants were divided into expert drivers, novice drivers, and passengers. Expert drivers (10 males, aged 35-48) had 12-28 years of experience and annual mileage ranging from under 20,000 km to over 100,000 km. Novice drivers (11 males, 3 females, aged 20-25) had 1-4 years of experience with mileage up to 40,000 km. The passenger group (48 people, aged 20-50) included 32 licensed individuals with limited driving experience and 16 without a license.

We first arranged for each driver to drive an urban road and two passengers to ride this vehicle, which covered 13 different driving conditions and a total distance of 5.7 kilometers. We installed a 360-degree panoramic camera on top of the vehicle to record the vehicle's surroundings in real time and captured the driver's behavior through multiple motion cameras inside the vehicle, which enabled us to accurately review the entire driving maneuvers and performance in the follow-up interviews. In addition, we recorded the vehicle's real-time CAN bus data to capture the vehicle's driving condition, including throttle, brakes, steers, etc.

During the interviews, we used the collected videos to recreate the driving process that the drivers and passengers had just experienced. For each driving context and action, participants were asked to verbalize their thought processes during the interview (while watching the video) and to consider how they evaluate driving behavior. Drivers shared their driving preferences and assessed whether driving behaviors were in alignment with their daily habits by looking back at the videos and reflecting on the logic behind the decisions and actions made while driving and the reasons for the differences. Passengers, for their part, reviewed comfortable and uncomfortable driving segments and expressed their feelings and riding habits under different driving conditions. 

These interview data constitute our original dataset, focusing on the driver's actions, decisions, and mindset to achieve the driving and the passenger's perception, expectation, and evaluation of the riding experience.

\subsection{Construct Driving-Evaluation Dataset}

To bridge the gap in driving-specific domain knowledge currently lacking in LLMs, we extracted key messages from the interview data and constructed a knowledge database of languages in the field of autonomous driving.

Firstly, we refined the interview data to ensure both accuracy and consistency. Because of the prevalence of colloquial expressions in the raw interviews, directly utilizing this unprocessed data could potentially hinder subsequent analysis and database development. To address this issue, we employed GPT-4 as a tool to transform the informal spoken content into more formal and concise textual data. This process not only enhances the readability of the dataset but also mitigates errors caused by linguistic redundancy, thereby ensuring a higher level of consistency and structure for further data processing stages.

Next, we classified and processed the refined data based on the distinct roles of drivers and passengers, aiming to analyze and understand the divergent knowledge and evaluations of driving behavior between these two groups. The driver data was processed into knowledge units according to five key elements: ``Context", which encapsulates the driving environment and traffic interactions; ``Driver Mindset", reflecting the driver's cognitive state and motivations; ``Driving Decision", which elucidates the reasoning behind the driver’s choices and actions in specific contexts; ``Driver Action", denoting the concrete operational behaviors executed by the driver; and ``Driver Evaluation", summarizing the driver’s assessment and reflection on their driving performance.

Passenger data units followed a different analytical framework: ``Context", describing the driving environment from the passenger's perspective; ``Passenger Mindset", which identifies the passenger’s core beliefs, concerns, and emotional state during the journey; ``Expectations", referring to the passenger’s anticipated behavior from the driver; ``Passenger Perception", capturing the passenger's actual experiences and reactions to the drive; and ``Passenger Evaluation", reflecting the passenger’s overall assessment of the driving experience.

By employing this structured categorization, we can effectively capture the distinct perspectives and cognitive processes of both drivers and passengers, providing a more nuanced and multi-dimensional foundation for building a knowledge database.

Finally, after processing the data, we convert it into JSON format to build an external knowledge base, including 700 Driver Knowledge Units and 760 Passenger Knowledge Units. This knowledge base enhances our system's evaluation using the Retrieval-Augmented Generation (RAG) mechanism. Providing a structured driving-related knowledge dataset, allows LLMs to generate more accurate and well-informed assessments, improving their reasoning and evaluation in this field.

\section{DRIVING EVALUATION FRAMEWORK}
%Evaluating comprehensive driving performance is fraught with challenges, primarily due to the intricate nature of real-world driving scenarios. Existing evaluation methods typically focus on isolated aspects of the driving experience, often relying on quantitative metrics that do not capture the full scope of driving behavior or contextual understanding. This narrow approach can lead to an incomplete assessment of performance, missing critical factors such as environmental influences, human perception, and decision-making intricacies. To overcome these limitations, our LLM-based autonomous driving evaluation system with the structure shown in Fig,~\ref{fig:framework} is designed to extract and process both quantitative and contextual information from driving data. and gives a comprehensive evaluation from Safety evaluation, three-step Intelligence Evaluation to Comfort Evaluation. This structured methodology ensures that evaluations are not only accurate but also reflective of the nuanced driving environments encountered in practice.
To address the limitations of traditional driving performance evaluation methods, which often focus on isolated quantitative metrics and overlook the complexities of real-world scenarios, contextual nuances, and decision-making intricacies, we propose an LLM-based autonomous driving evaluation system (Fig.~\ref{fig:framework}). This system processes both quantitative and contextual driving data, providing a comprehensive assessment encompassing safety, intelligence, and comfort. Our approach ensures that evaluations are accurate and reflective of nuanced driving environments encountered in practice.

\subsection{From driving data to understandable driving context}

In autonomous driving, interpreting vast amounts of complex real-world data is crucial for safe and efficient algorithm operation. To enhance evaluation stability, we transform raw driving data into structured driving contexts. Our system categorizes this data into five types:

\begin{itemize}
\item Driving Scenarios: Information about the route setup
\item Weather Conditions: Factors influencing driving (e.g., fog density affecting visibility).
\item Ego-vehicle CAN BUS Data: Real-time vehicle metrics.
\item Surrounding Vehicles and Pedestrians: Data on nearby vehicles (two closest in the same lane, closest in adjacent lanes, and closest opposing traffic), pedestrians, and special vehicles like firetrucks.
\item Stationary Actors: Relevant objects like traffic lights and signs facing the ego-vehicle.
\end{itemize}

Our approach summarizes the route and weather conditions, followed by collision information and CAN BUS data. We then format detailed descriptions of surrounding vehicles and driving-relevant objects.

\subsection{LLM-powered Driving Evaluation Framework}
Building upon our driving evaluation dataset, we construct a comprehensive evaluation framework encompassing three primary dimensions: Safety Evaluation, Intelligence Evaluation, and Comfort Evaluation: %. This multi-dimensional approach was designed to provide a holistic understanding of the driving scenario.

\subsubsection{Safety evaluation}
The safety dimension, being the most mature and quantifiable aspect of traditional evaluations, is a key component of our framework. Our safety evaluation focuses on quantifying various risk factors in the driving environment and assessing the driver's adherence to traffic regulations and ability to navigate potential hazards. This assessment encompasses common autonomous driving evaluation metrics, including collision avoidance, appropriate handling of traffic signs, and adherence to traffic lights.
%The safety evaluation dimension focused on quantifying the various risk factors present in the driving environment, as well as assessing the driver's adherence to traffic regulations and ability to navigate potential hazards. This evaluation of safety covered the common aspects used in autonomous driving evaluation metrics, which included analyzing factors such as avoiding collisions with other road users, appropriately handling traffic signs, and adhering to traffic lights. 

\subsubsection{Intelligence Evaluation}
The intelligence dimension represents the most critical metric extracted from our Driving-Evaluation dataset. In our framework, this evaluation assesses the driver's adaptive behaviors, decision-making processes, and strategic planning in complex traffic situations. We employ a Chain-of-Thought (CoT) methodology, encompassing three hierarchical driving levels: Operational, Tactical, and Strategic.
%The intelligence evaluation dimension was concerned with assessing the driver's adaptive behaviors, decision-making processes, and strategic planning in response to complex traffic situations. This evaluation is executed through a Chain-of-Thought (CoT) methodology, which encompasses three hierarchical levels of driving: Operational Level, Tactical Level, and Strategic Level.

Our hierarchical evaluation implements cascading reasoning - operational outputs inform tactical analysis, which subsequently feeds strategic assessment. At each level, the LLM dynamically retrieves relevant cases from our structured knowledge base via RAG, providing domain-specific knowledge.

\paragraph{Operation Level}
The operational level evaluates the driver's actions with the driving context, focusing on the driving operation performance, and evaluating whether the vehicle is well-controlled through the testing criteria:

\begin{itemize}
    \item Driving Stability. Vehicle's ability to maintain stable control, focusing on steady speed and consistent direction without unnecessary adjustments or swerves. 
    \item Operation Fluidity. Smoothness and coherence of the vehicle's maneuvers, such as lane change and overtake. 
    \item Anomaly Handling. Vehicle's capacity to detect anomalies, make timely decisions, and navigate safely through scenarios like road constructions or accidents.
    \item Reaction Speed. Reaction time of the vehicle when confronted with sudden events.
\end{itemize}

\paragraph{Tactical Level}
Following the operational assessment, both the driving context and the operational evaluations will be passed to the tactical level evaluation. The tactical level evaluation delves into the decision-making processes of the vehicle, encompassing:
\begin{itemize}
    \item Social Intelligence. The vehicle’s ability to communicate intentions and follow social driving norms, such as turning the vehicle heading to the target direction, and providing great predictability for other road users.
    \item Coping Complex Scenarios. The vehicle's problem-solving abilities in challenging environments like urban intersections or pedestrian-heavy areas, reflect the vehicle's adaptability to make quick and appropriate decisions to navigate safely and highly efficiently. 
    \item Strategic Competence. Vehicle's ability to anticipate other road users' actions. Address the ability to handle the situation when multiple road users have conflicting interests, and make optimal choices to avoid conflicts while maximizing passing efficiency.
    \item Covert Hazard Prediction. Vehicle’s proficiency in anticipating hidden risks, focusing on the vehicle's predictive ability and preparedness for potential hazards.

    \item Decision Optimality. The consistency and quality of the vehicle's decision-making processes across various driving scenarios, assessing its ability to make optimal choices in diverse situations.
\end{itemize}

\paragraph{Strategic Level}
Enhanced by the result from the previous two levels and driving context, the Strategic Level assesses the vehicle's higher-level understanding of the driving environment. This level of evaluation focuses on the mindset of the vehicle and analyzes its strategy of driving in these aspects:
\begin{itemize}
    \item Macro-level Transportation Vision. Understand and predict broader traffic patterns, assessing its capacity to contribute to overall traffic flow optimization.
    \item Driving Style. Risk tolerance and driving style between aggressive and cautious, evaluating how it balances safety and efficiency in various driving contexts.
    \item Vulnerable Road User Consideration. Adherence to ethical driving principles and social norms, assessing the ability to prioritize vulnerable road users like bikes.
    \item Passengers Consideration. The ability to provide a smooth and pleasant driving experience, reflects its capacity to balance performance with passenger comfort.
    \item Environmental Consciousness. The vehicle's ability to optimize its driving patterns for energy efficiency, reflects its contribution to environmental sustainability.
    \item Proactive Safety Driving. Ability to anticipate and prevent potentially dangerous situations, assessing its proactive approach to safety.
\end{itemize}

\subsubsection{Comfort Evaluation}
The comfort dimension, a well-established metric in traditional vehicle assessment, is integrated into our framework by extracting passenger evaluation logic from the Driving Evaluation dataset. Focusing on the human experience within the vehicle. It recognizes that transportation aims not only for mobility but also for user comfort and perceived safety.
We address two key perceptions:

\begin{itemize}
\item Passenger Perception of Comfort: Evaluates comfort based on vehicle actions (speed, throttle, braking, steering) and efforts to ensure a comfortable ride at tactical and strategic levels.
\item Passenger Perception of Safety: Passenger feeling about vehicle actions, particularly during sudden events.
\end{itemize}

% The Comfort Evaluation of potential passengers is a crucial component of assessing the overall performance of driving. This dimension focuses on the human experience within the vehicle, recognizing that the ultimate goal of transportation extends beyond mere mobility to encompass user comfort and perceived safety. By examining these subjective yet critical aspects, we can gain valuable insights into the vehicle's ability to provide a satisfactory and trustworthy travel experience.
% To achieve a comprehensive evaluation of passengers' experience, two different perceptions are addressed with the help of results from all three levels of intelligence evaluation:
% \begin{itemize}
%     \item Passenger Perception of Comfort. This criterion evaluates the comfort level based on the actions of the vehicle, considering factors such as speed, throttle, braking, and steering, and the effort to create a comfortable ride at the tactical and strategic level.
%     \item Passenger Perception of Safety. This criterion evaluates passenger feedback on the actions of the vehicle, especially when encountering sudden events.
% \end{itemize}

\subsubsection{Evaluation Conclusion}

After the sequential evaluations across these levels and dimensions, a summary of the driving behavior is produced. This assessment identifies areas for improvement and culminates with an overall driving score, providing a holistic view of the vehicle's performance. For the complete evaluation prompt used in this framework, please refer to the link provided in the abstract.

%The evaluation framework adopts uniform weighting schemes: (1) Safety, Intelligence, and Comfort dimensions share equal weights, and (2) within Intelligence evaluation, operational, tactical, and strategic sub-levels also maintain equal weighting. This baseline configuration demonstrates the framework's core performance, while the modular design allows customized weight allocation through prompt engineering%

%Exploring adaptive weight allocation strategies based on driving scenarios through dynamic prompt generation (Add to discussion?)%

\section{EXPERIMENT}
\subsection{Experiment Environment Setup}
In our experiment, we build the CARLA Autonomous Driving Leaderboard 2.0 platform on a Lenovo ThinkPad T15g Gen 1 Laptop with Ubuntu 22.04.4 LTS operation system, as the driving simulator for data collection. The leaderboard, different from the original CARLA, provides a series of predefined routes, which include 6 types in total 24 traffic scenarios. The predefined routes cover urban routes, freeways, residential regions, rural settings, and a variety of common weather conditions. In the experiment, we leverage OpenAI’s GPT-4o APIs for performing our driving evaluation framework.

\subsection{Experiment Driving Data Collection}
To comprehensively evaluate the capabilities of our assessment framework, we generated a diverse collection of driving data using CARLA Leaderboard. We collect all driving data by manual driving for better controllability. In order to retain the diversity of driving situations and weather conditions provided by predefined routes, cover as many different scenarios as possible, and avoid the problem of drowsy driving, we split the predefined driving routes into shorter routes by every five driving scenarios. By computing the combination of 8 short routes that cover most combinations of different driving scenarios, driving situations, and weather, we end up covering all types of driving scenarios, driving situations of urban routes, freeways, residential regions, and weather of daylight, rain, and fog. The data were collected once per second. 
Our experimental conditions combined two driving styles (Cautious/Aggressive) and two performance levels (Good/Bad): 
\textit{Cautious-Good;  Cautious-Bad; Aggressive-Good; Aggressive-Bad}.
During the experiment, the four types of driving conditions are collected according to the guidelines, part of the criteria provided below:

%For driving performance, we covered two driving styles and two driving skill levels, resulting in a total of four experimental conditions as follows.
%\begin{itemize}
%    \item Condition 1: Cautious driving with Good performance
%    \item Condition 2: Cautious driving with Bad performance
%    \item Condition 3: Aggressive driving with Good performance
%    \item Condition 4: Aggressive driving with Bad performance
%\end{itemize}

%cautious driving with good performance on the test route, cautious driving with poor performance on the test route, aggressive driving with good performance on the test route, and aggressive driving with poor performance on the test route. 

\subsubsection{Criteria for good and bad performance}

%Common indicators of \textbf{good driving performance} include: decisive operation, adherence to social norms and traffic rules, maintaining speed consistency, decelerating using the vehicle's natural speed reduction, high overall traffic efficiency, taking defensive measures in response to other vehicles' actions in advance, and the absence of collisions, among others.
\textbf{Good driving performance} include: adherence to traffic rules, maintaining speed consistency and the absence of collisions, etc.
%Common indicators of\textbf{ bad driving performance} include: frequent collisions with other vehicles, low efficiency, poor speed consistency, excessive acceleration and deceleration, frequent hard braking, sudden steering, lack of consideration for other road users, excessively reckless driving (common in aggressive styles), excessive hesitation (common in cautious styles), etc.
\textbf{Bad driving performance} include: frequent collisions, poor speed consistency, excessively reckless driving (in aggressive styles), excessive hesitation (in cautious styles), etc.
%(alternating between fast and slow speeds)

%Based on these criteria, the overall performance on each route was subjectively evaluated by the experimenters and categorized as either ``good" or ``bad."

\subsubsection{Criteria for aggressive and cautious driving styles}
%Driving styles were classified into aggressive and cautious. The criteria for distinction mainly considered the fundamental characteristics of each style.

%The \textbf{aggressive driving style} typically includes but is not limited to the following traits: strong acceleration and deceleration, frequent hard braking, minimal use of the vehicle’s natural speed reduction to decelerate, and maintaining small distances from other vehicles. 
\textbf{Aggressive driving style}: Strong acceleration and deceleration, minimal use of the vehicle’s natural speed reduction to decelerate, and maintaining small following distances, etc.
%The advantages may include: quick reactions and ability to brake in time to avoid collisions; efficiently overtaking slow-moving bicycles or stationary vehicles/obstacles. Disadvantages may include: frequent failure to brake in time, leading to collisions with other vehicles, and maintaining an excessively short distance from the vehicle ahead.
\textbf{Cautious driving style}: Smoother acceleration and deceleration, slowing down to observe during complex road conditions, and maintaining a greater following distance.
%The advantages may include: keeping an appropriate distance from other vehicles, utilizing the vehicle’s natural speed reduction to decelerate, and smooth acceleration and deceleration. Disadvantages may include: being overly cautious, such as following slow-moving bicycles instead of overtaking; maintaining a large following distance but frequently braking prematurely when far from potential risks; hesitating at intersections and lacking decisiveness when making turns or overtaking, which could result in collisions.

%Due to the limited capabilities of the sample autonomous driving algorithms provided by CARLA, we developed a \textbf{driving style guide} based on the previously mentioned characteristics of driving styles. The experimenters then manually drove according to the guide in order to approximate the desired driving styles and effects as closely as possible.

These class-specific data also enable basic hallucination detection when LLM evaluations violate expected value ranges, for example, a high score in Proactive Safety Driving for Aggressive driving data is rare.
%aggressive driving style was simulated and recorded by increasing the acceleration curve slope, raising the speed limit, and manually mimicking aggressive driving. Similarly, cautious driving style was simulated and recorded by decreasing the acceleration curve slope, lowering the speed limit, and manually mimicking cautious driving.

\subsection{System Evaluation}
In our system setup, the three dimensions and three levels share equal weight during the corresponding evaluation, demonstrating the framework's baseline performance. The weighting factors are parameterizable through prompts to accommodate domain-specific requirements. 

To understand the ability to evaluate driving conditions, we examine the accuracy of our evaluation results in the aspects of identifying between two driving styles, and distinguishing between the two driving performances. By randomly selecting one scenario from each of the eight test routes, and performing evaluations on the four driving conditions, the system correctly identifies 30 out of 32 scenarios in driving style, accounting for an accuracy of 93.75\%. In the driving performance evaluation, the result correctly identifies 26 out of the 32 scenarios, accounting for an accuracy of 81.25\%. %Furthermore, a Spearman correlation analysis between our system and CARLA's scoring system yielded a strong and statistically significant correlation ($\rho = 0.561$, $p < 0.001$), demonstrating a consistent evaluation trend between both systems. This indicates that our evaluation results align well with CARLA's assessments, further validating the reliability of our scoring approach.%
%modification

Furthermore, we performed a Spearman correlation analysis between the total intelligence scores assigned by our system and the driving scores of each scenario computed by the CARLA leaderboard. The analysis revealed a moderate and statistically significant correlation ($\rho = 0.561$, $p < 0.001$), suggesting a consistent evaluation trend between the two scoring systems. This finding indicates that our scoring method aligns reasonably well with CARLA's established assessment criteria, further supporting the validity and reliability of our evaluation approach.

\subsection{Human Evaluation}
\subsubsection{Procedure}
Targeting to understand whether the human evaluation is consistent with our LLM-generated evaluation, we designed a survey questionnaire to collect human drivers' evaluations of the performance of our evaluation framework, which was presented to participants in the questionnaire through video clips of the driving footage.
We initially collected basic information from participants. %(e.g. age, gender, whether holding a driving license, driving experience, and driving style) from participants.
%Each group of video clips will appear in a random order, accompanied by five rating questions and one reasoning question. Three of the rating questions each contain a different level of intelligence evaluation, and the remaining two contain the comfort level evaluation and overall evaluation result. These rating questions require participants to rate the level of acceptance of the evaluation results from 0 to 10; the higher score presents a higher level of acceptance. A reasoning question followed, requiring the participants to explain their rating.
The participants evaluated the results of the framework with the corresponding video clip in randomized order. Each trial was accompanied by five rating questions and one reasoning question. Three of the five questions were designed to assess the degree of agreement on different levels of intelligence evaluation, while the remaining two addressed comfort evaluation and the overall result of the evaluation. These rating questions require participants to rate the level of agreement of the evaluation results from 0 to 10. The basic format of the rating questions in the questionnaire is as follows.
%the driving styles in the videos according to their level of riskiness (a smaller number indicates more risk) and a reason question for explaining their rankings. (all from 0 to 10) and a reason question attached below each clip.

\textit{
``Based on the descriptions of the following five aspects, please rate your degree of agreement with the generated content. A score of 10 indicates complete agreement, while a score of 0 indicates complete disagreement. Lower scores represent lower levels of agreement."}

The reasoning question, which requires the participants to explain their ratings, appears as the final question for each experiment trial.

%\begin{tcolorbox}[colback=gray!10, colframe=black, title= Rating Question Format]
%\textbf{Please first watch the provided video demonstration, which displays the performance of an autonomous driving system in a specific environment.}

%Based on descriptions of the following five aspects, please rate your degree of agreement with the generated content. A score of 10 indicates complete agreement, while a score of 0 indicates complete disagreement. Lower scores represent lower levels of agreement.
%\end{tcolorbox}

Additionally, to filter out carelessly completed questionnaires, we set a minimum answering time and included trap questions in the questionnaire, which required participants to select a particular option.

\subsubsection{Participants}
Our research team recruited participants through social media platforms, attracting over 60 individuals with a compensation of approximately \$2.08 per valid questionnaire. After verification, 33 participants with valid driving licenses were included in the final analysis. The sample comprised 17 male (52.22\%) and 16 female (47.78\%), with ages ranging from 19 to 54 years.
Participants' driving experience varied widely. The majority had relatively low driving experience, with 33.3\% having 0-2 years and 39.4\% having 3-5 years of experience. More experienced drivers were also represented, with 12.1\% having 6-10 years, 9.1\% having 11-20 years, and 6.1\% having over 21 years of driving experience.
An interesting distribution emerged regarding self-reported driving styles. The majority (66.7\%) of participants described their daily driving style as cautious. Nearly a third (30.3\%) reported a neutral or situational driving style, adapting their approach based on circumstances. Only one participant identified the daily driving style as aggressive. 

\subsubsection{Results}
\begin{figure}
    \centering
    \includegraphics[width=1\linewidth]{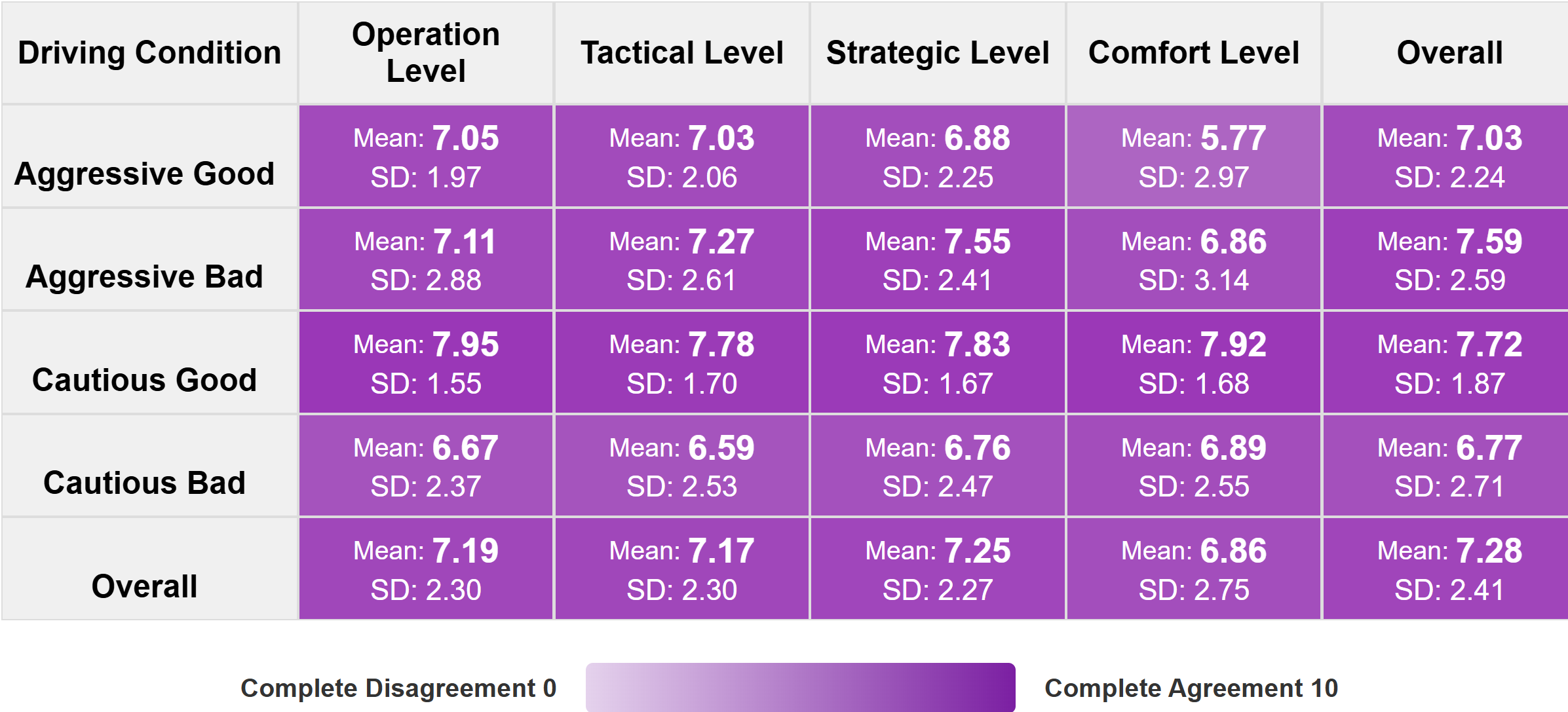}
    \caption{Human Agreement Score on System Evaluation}
    \label{fig:heatmap}
\end{figure}

In the evaluation of our LLM-powered framework, human drivers assessed various dimensions, including operational, tactical, strategic, comfort, and overall performance. Each dimension was rated on a scale from 0 to 10, with the results indicating a generally positive reception. The summary of human evaluation results is shown in Fig.~\ref{fig:heatmap}.

%\subsection{Result based on driving condition category}
%The human evaluation requires repeated measures to prove the generated result
\subsection{Findings}

Overall, we found that human drivers provided favorable comprehensive evaluations of our system (7.28 out of 10). Notably, the acceptance levels for the three intelligence-related levels were remarkably consistent (7.19, 7.17, 7.25 out of 10), validating the rationality of our chain-of-thought design and its alignment with human driver evaluations. 

Participants appreciated the objectivity and thoroughness of the safety assessment, as evidenced by the following comments:
\textit{'I concur with the low safety level assessment. The system's evaluation is highly objective and comprehensive.' (aggressive, bad condition)}
\textit{'I agree with the identified issues in autonomous driving operations continuity. The system's evaluation is exceptionally thorough, analyzing perspectives of both the autonomous vehicle and other vehicles.' (cautious, bad condition);}
However, the comfort evaluation received lower acceptance (6.86 out of 10). One participant remarked:
\textit{'I disagree with the comfort assessment. The vehicle's sudden acceleration during turns presents safety risks and low comfort. The system should strengthen its passenger comfort evaluation criteria.' (aggressive, bad condition)}
Comfort perception varied considerably among participants, with some finding the comfort evaluation appropriate for the same video clip.

Among the four different driving conditions, our Cautious Good condition currently receives the highest level of acceptance (7.72 out of 10). In contrast, both the Cautious Bad and Aggressive Good conditions currently have relatively lower levels of acceptance (6.77 and 7.03 out of 10, respectively). This indicates that, comparatively, participants exhibited a lower degree of approval for the LLM's ratings in these two driving conditions. This disparity may indicate that driving styles characterized as cautious or aggressive are inherently coupled with driving performance (good or bad), resulting in greater difficulty in achieving consensus in evaluations for these conditions~\cite{yang2024driving}.
The human validation study also serves as a critical hallucination detection mechanism, with low acceptance showing possible hallucination happening or incorrect evaluation occurring.

\section{CONCLUSIONS AND FUTURE WORK}

This paper introduces a novel evaluation framework for assessing driving intelligence in complex traffic scenarios, addressing a critical gap in autonomous driving research. 

Our evaluations demonstrate the framework's potential to provide nuanced, context-aware assessments of driving behavior intelligence, contributing significant insights to the field. However, limitations include reliance on simulated environments, potential biases in the natural language dataset, the possible inconsistency in manual driving data collection, and challenges with rare scenarios. The real-world driving data accounts for different driving levels and road conditions, but safety-critical scenarios, such as accidents or extreme situations, are rare. For example, in the two most recent years (2022 and 2023), in 22 countries, the number of fatalities rarely exceeded 8 per billion vehicle-kilometers, with a median of 4-5 cases\cite{ITF2024}. This suggests that safety-critical scenarios may not be adequately represented. Future work could address this limitation by integrating high-quality simulation results into the knowledge dataset.  Further exploration could also consider the framework's adaptability to diverse driving cultures, integration into real-world testing protocols, multi-modal data incorporation, edge case analysis, and the development of explainable AI.
%Future work could address this limitation by incorporating simulated safety-critical scenarios, which would help to enhance the robustness of the framework. 
%One notable limitation is that the real-world driving data used may lack coverage of safety-critical scenarios, such as accidents or extreme situations. While real-world driving data often accounts for different driving levels and road conditions, collision events are rare. For example, in the two most recent years (2022 and 2023), in 22 countries, the number of fatalities rarely exceeded 8 per billion vehicle-kilometers, with a median of 4-5 cases\cite{ITF2024}. This suggests that safety-critical scenarios, though important for robustness, may not be adequately represented. Future work could address this limitation by incorporating simulated safety-critical scenarios, which would help to enhance the robustness of the framework. Further exploration could also consider the framework's adaptability to diverse driving cultures, integration into real-world testing protocols, multi-modal data incorporation, edge case analysis, and the development of explainable AI.
Additionally, conducting longitudinal studies, evaluating ethical decision-making, and exploring human-AI collaboration in semi-autonomous vehicles present promising research directions. %Investigate the multi-agent interaction, and discover the possibility of cooperation between autonomous vehicles and the evaluation agent, improving overall traffic safety and efficiency.  
These efforts aim to enhance the framework's applicability and effectiveness in advancing autonomous driving technology, ultimately leading to more sophisticated, human-like autonomous driving agents and comprehensive, human-centric evaluation systems.

\section*{ACKNOWLEDGMENT}

This research is supported by the National Key R\&D Program of China (2022YFB4300300), Beijing Municipal Science and Technology Project (Nos. Z231100010323005), and Beijing Natural Science Foundation (Grant No.L233033).

%%%%%%%%
\bibliographystyle{IEEEtran}
\bibliography{IEEEfull,root,bib/ADEvaluation,bib/LLMinDriving}

\end{document}